%% file: emnlp2017.tex
\documentclass[11pt,letterpaper]{article}
\usepackage{emnlp2017}
\usepackage{times}
\usepackage{latexsym}
\usepackage{paralist}
\usepackage{amsmath}
\usepackage{mathrsfs,amssymb}
\usepackage{enumitem}
\usepackage{color}
\usepackage{graphicx}
\graphicspath{figures}
\usepackage{caption} 
\usepackage[labelformat=simple]{subcaption}

\usepackage{xspace}
\usepackage{array}
\usepackage[ruled,vlined]{algorithm2e}

\emnlpfinalcopy

\def\emnlppaperid{1470}

\newcommand{\alg}{RBA\xspace}

\title{Men Also Like Shopping:\\Reducing Gender Bias Amplification using Corpus-level Constraints}

\author{Jieyu Zhao$^\S$ \qquad
 Tianlu Wang$^\S$ \qquad 
  Mark Yatskar$^\ddag$   \\
{\bf Vicente Ordonez$^\S$ \qquad
 Kai-Wei Chang$^\S$}
\\
  $^\S$University of Virginia \\ 
  \{jz4fu, tw8cb, vicente, kc2wc\}@virginia.edu
  \\ $^\ddag$University of Washington \\  my89@cs.washington.edu
}

\hypersetup{draft}
\date{}
\begin{document}

\maketitle

\input{sections/abstract.tex}
\newcolumntype{C}[1]{>{\centering\arraybackslash}p{#1}}

\section{Introduction}
\label{sec:introduction}
\input{sections/introduction.tex}

\section{Related Work}
\label{sec:related_work}
\input{sections/related_work.tex}

\section{Visualizing and Quantifying Biases}
\label{sec:general}
\input{sections/method_quantify_bias.tex}

\section{Calibration Algorithm}
\label{sec:calibration}
\input{sections/method_calibrate.tex}

\section{Experimental Setup}
\label{sec:setup}
\input{sections/experimental_setup.tex}

\section{Bias Analysis} 
\label{sec:results_bias}
\input{sections/results_bias.tex}

\section{Calibration Results}
\label{sec:results_calibration}
\input{sections/results_calibration.tex}

\section{Conclusion}
\label{sec:conclusion}
\input{sections/conclusion.tex}

%\section*{Acknowledgments}
%\section{Appendix}
%\label{sec:appendix}
%\input{sections/appendix.tex}

\bibliography{emnlp2017,cited,ccg}
\bibliographystyle{emnlp_natbib}
\end{document}

%% file: sections/abstract.tex
\begin{abstract}
Language is increasingly being used to define rich visual recognition problems
%, such image captioning or visual semantic role labeling, 
with supporting image collections sourced from the web.
Structured prediction models are used in these tasks to take advantage of correlations between co-occurring labels and visual input but risk inadvertently encoding social biases found in web corpora.  
%but also also run risk of encoding biased correlation   
%Large scale datasets supporting such efforts are often sourced from the web, and encode significant social 
% into such visual recognition systems.
In this work, we study data and models associated with multilabel object classification and visual semantic role labeling.
%, a structured activity recognition formalism.
%based on semantic role labeling and image captioning(?).
We find that (a) datasets for these tasks contain significant gender bias and (b) models trained on these datasets further amplify existing bias.
For example, the activity \texttt{cooking} is over 33\% more likely to involve females than males in a training set, and a trained model further amplifies the disparity to 68\% at test time.
We propose to inject corpus-level constraints for calibrating existing structured prediction models and design an algorithm based on Lagrangian relaxation for collective inference.
Our method results in almost no performance loss for the underlying recognition task but decreases the magnitude of bias amplification by 47.5\% and 40.5\% for multilabel classification and visual semantic role labeling, respectively.

 %Vision-and-language models have been designed and applied to tasks such as image captioning and visual semantic role labeling. Despite their promising performance, they run the risk of amplifying implicit human biases. For example, an image captioning system produces ``a woman is cooking'' even though the image clearly presents ``a man cooking.'' This is likely because implicit biases present in the training data are captured by the model.
 %In this paper, we propose a general framework to visualize and quantify the biases in a vision-and-language model. We then apply the framework in analyzing an image captioning system and a visual semantic role labeling system. We show that both systems significantly amplify the implicit associations on the training data, generating biased outputs.
 
\end{abstract}

%% file: sections/introduction.tex
\begin{figure*}
\centering
    \includegraphics[width = \linewidth]{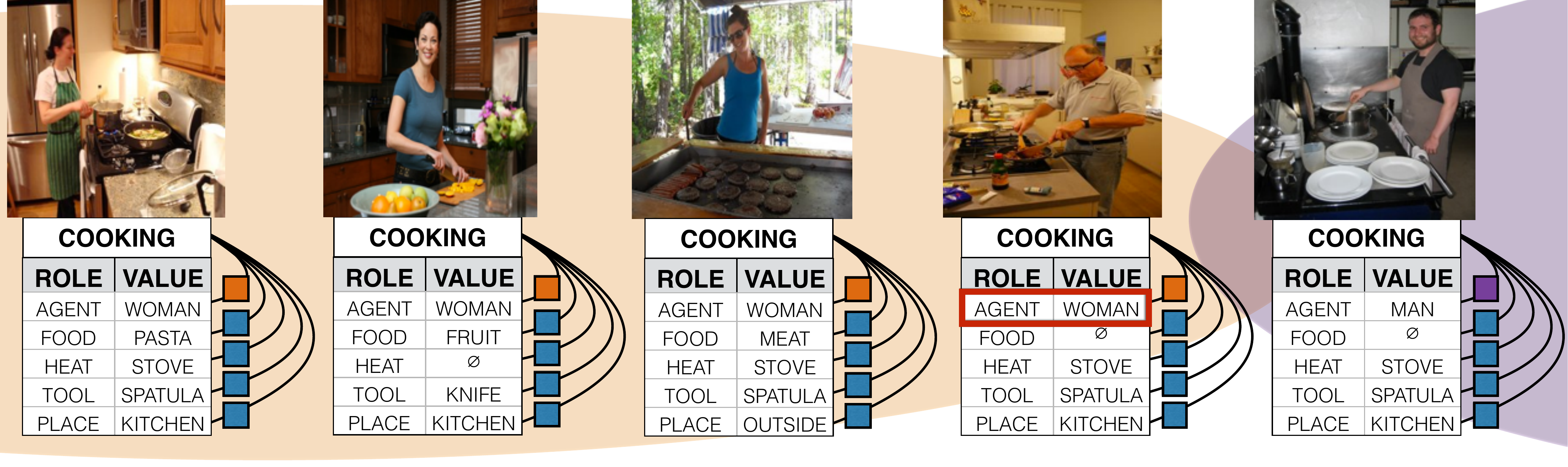}
\caption{ 
Five example images from the imSitu visual semantic role labeling (vSRL) dataset.
Each image is paired with a table describing a situation: the verb, \texttt{cooking}, its semantic roles, i.e \texttt{agent}, and noun values filling that role, i.e. \texttt{woman}.
In the imSitu training set, 33\% of \texttt{cooking} images have \texttt{man} in the \texttt{agent} role while the rest have \texttt{woman}.
After training a Conditional Random Field (CRF), bias is amplified: \texttt{man} fills 16\% of \texttt{agent} roles in \texttt{cooking} images.
To reduce this bias amplification our calibration method adjusts weights of CRF potentials associated with biased predictions.
After applying our methods, \texttt{man} appears in the \texttt{agent} role of 20\% of \texttt{cooking} images, reducing the bias amplification by 25\%, while keeping the CRF vSRL performance unchanged.}

\label{fig:cooking}
\end{figure*}
Visual recognition tasks involving language, such as captioning~\cite{vinyals2015show}, visual question answering~\cite{antol2015vqa}, and visual semantic role labeling~\cite{yatskar2016situation}, have emerged as avenues for expanding the diversity of information that can be recovered from images.
These tasks aim at extracting rich semantics from images and require large quantities of labeled data, predominantly retrieved from the web.
Methods often combine structured prediction and deep learning to model correlations between labels and images to make judgments that otherwise would have weak visual support.
%To model language structure and the correlation between language and vision, existing approaches combine structured prediction with deep learning. 
%in modeling the interdependencies among input and output variables.
For example, in the first image of Figure~\ref{fig:cooking}, it is possible to predict a \texttt{spatula} by considering that it is a common tool used for the activity \texttt{cooking}.
Yet such methods run the risk of discovering and exploiting societal biases present in the underlying web corpora.
%Such methods are adept at leveraging correlations between input and output variables but run the risk discovering and exploiting social biases, and, as we show later, even amplify these biases.
Without properly quantifying and reducing the reliance on such correlations, broad adoption of these models can have the inadvertent effect of magnifying stereotypes.

In this paper, we develop a general framework for quantifying bias and
study two concrete tasks, visual semantic role labeling (vSRL) and multilabel object classification (MLC). 
In vSRL, we use the imSitu formalism \cite{yatskar2016situation,yatskar2016commonly}, where the goal is to predict activities, objects and the roles those objects play within an activity.
For MLC, we use MS-COCO~\cite{lin2014microsoft,chen2015microsoft}, a recognition task covering 80 object classes.
 We use gender bias as a running example and show that both supporting datasets for these tasks are biased with respect to a gender binary\footnote{To simplify our analysis, we only consider a gender binary as perceived by annotators in the datasets. We recognize that a more fine-grained analysis would be needed for deployment in a production system. Also, note that the proposed approach can be applied to other NLP tasks and other variables such as identification with a racial or ethnic group.}.
 
Our analysis reveals that over 45\% and 37\% of verbs and objects, respectively, exhibit bias toward a gender greater than 2:1. 
For example, as seen in Figure~\ref{fig:cooking}, the \texttt{cooking} activity in imSitu is a heavily biased verb.
Furthermore, we show that after training state-of-the-art structured predictors, models amplify the existing bias, by 5.0\% for vSRL, and 3.6\% in MLC.

To mitigate the role of bias amplification when training models on biased corpora, we propose a novel constrained inference framework, called RBA, for {\bf R}educing {\bf B}ias {\bf A}mplification in predictions. Our method introduces corpus-level constraints so that gender indicators co-occur no more often together with elements of the prediction task than in the original training distribution. 
For example, as seen in Figure~\ref{fig:cooking}, we would like noun \texttt{man} to occur in the \texttt{agent} role of the \texttt{cooking} as often as it occurs in the imSitu training set when evaluating on a development set.
We combine our calibration constraint with the original structured predictor and use Lagrangian relaxation~\cite{korte08,Rush2012} to reweigh bias creating factors in the original model.
%Our method makes few assumptions about the underlying predictor beyond access to an efficient inference algorithm and is ?? some other positive statement ??. 

We evaluate our calibration method on imSitu vSRL and COCO MLC and find that in both instances, our models substantially reduce bias amplification. 
For vSRL, we reduce the average magnitude of bias amplification by 40.5\%. 
For MLC, we are able to reduce the average magnitude of bias amplification by 47.5\%. 
Overall, our calibration methods do not affect the performance of the underlying visual system, while substantially reducing the reliance of the system on socially biased correlations\footnote{Code and data are available at \url{https://github.com/uclanlp/reducingbias}}.

%% file: sections/related_work.tex
As intelligence systems start playing important roles in our daily life, 
ethics in artificial intelligence research has attracted significant interest.
It is known that big-data technologies sometimes inadvertently worsen 
discrimination due to implicit biases in data~\cite{PPMHZ14}. 
Such issues have been demonstrated in various learning systems, including
online advertisement systems~\cite{sweeney2013discrimination}, word embedding
models~\cite{bolukbasi2016man,CBN17}, online news~\cite{ross2011women}, 
web search~\cite{kay2015unequal}, and credit score~\cite{hardt2016equality}.
Data collection biases have been discussed in the context of creating image corpus~\cite{misra2016seeing,van2016stereotyping} and text corpus~\cite{gordon2013reporting,van2010extracting}. In contrast, we show that given a gender biased corpus, structured models such as conditional random fields, amplify the bias. 

The effect of the data imbalance can be easily detected and fixed when the prediction task is simple. 
For example, when classifying binary data with unbalanced labels (i.e., samples in the majority class dominate the dataset), a classifier trained exclusively to optimize accuracy learns to always predict the majority label, as the cost of making mistakes on samples in the minority class can be neglected. 
Various approaches have been proposed to make a ``fair'' binary classification~\cite{barocas2014big,dwork2012fairness,feldman2015certifying,zliobaite2015survey}.
For structured prediction tasks the effect is harder to quantify and we are the first to propose methods to reduce bias amplification in this context. 

Lagrangian relaxation and dual decomposition techniques have been widely used in NLP tasks~(e.g., \cite{sontag2011introduction,Rush2012,chang2011exact,peng2015dual}) for dealing with instance-level constraints. Similar techniques~\cite{ChangSuKe13,dalvi2015constrained} have been applied in handling  corpus-level constraints for semi-supervised  multilabel classification. 
In contrast to previous works aiming for improving accuracy performance, we incorporate corpus-level constraints for reducing gender bias.  

%% file: sections/method_quantify_bias.tex
Modern statistical learning approaches capture %explicit and implicit 
correlations among output variables in order to make coherent predictions. 
However, for real-world applications, some implicit correlations are not appropriate, especially if they are amplified. 
In this section, we present a general framework to analyze inherent biases learned and amplified by a prediction model. 

\paragraph{Identifying bias}
We consider that prediction problems involve several inter-dependent output variables $y_1, y_2, ... y_K$, which can be represented as a structure $y = \{y_1, y_2, ... y_K\} \in Y$. This is a common setting in NLP applications, including tagging, and parsing. For example, in the vSRL task, the output can be represented as a structured table as shown in  Fig \ref{fig:cooking}. Modern techniques often model the correlation between the sub-components in $y$ and make a joint prediction over them using a structured prediction model. More details will be provided in Section \ref{sec:calibration}.

We assume there is a subset of output variables $g\subseteq y, g\in G$ 
that reflects demographic attributes such as gender or race (e.g. 
$g\in G=\{\texttt{man}, \texttt{woman}\}$ is the agent), and there is another subset of the output 
$o\subseteq y, o\in O$ that are co-related with $g$ (e.g., $o$ is the activity present in an image, such as \texttt{cooking}). The goal is to identify the correlations that 
are potentially amplified by a learned model.

To achieve this, we define the bias score of a given output, $o$, with respect to a demographic variable, $g$, as:
%\begin{equation}
$$	b(o,g) = \frac{c(o, g)}{\sum_{g'\in G}c(o, g')},$$
    %\label{eq:b_ratio}
%\end{equation}
where $c(o,g)$ is the number of occurrences of $o$ and $g$ in a 
corpus.   
For example, to analyze how genders of agents and activities are co-related in 
vSRL, we define the gender bias toward \texttt{man} for each verb $b(verb, \texttt{man})$ as:
\begin{equation}
	 \frac{c(verb, \texttt{man})}{c(verb, \texttt{man}) + c(verb, \texttt{woman})}.
    \label{eq:gender_ratio}
\end{equation}
If $b(o,g) > 1/\|G\|$,  then $o$ is positively correlated with $g$ and may exhibit bias. 
%We test 
%the following null hypothesis in order to verify if the model is 
%``fair''. 
%\begin{hyp} \label{hyp:first}
%Let $Y^g=\{y^g\}$ and $Y^p=\{y^p\}$ be 
%\end{hyp}
%We first identify the attributes that are potentially biased. 
%potentially cause biased predictions and setup a hypothesis that the
%co-occurrences between these attributes are undesirably amplified by the model.  
%In this paper, we take gender biases occurred between ``man'' and ``woman'' as %example.
\paragraph{Evaluating bias amplification} 
To evaluate the degree of bias amplification, we propose to compare bias scores on the training set, $b^*(o,g)$, with bias scores on an unlabeled evaluation set of images $\tilde{b}(o,g)$ that has been annotated by a predictor. 
We assume that the evaluation set is identically distributed to the training set. Therefore, if $o$ is positively correlated with $g$ (i.e, ${b^*}(o,g) > 1/\|G\|$) and $\tilde{b}(o,g)$ is larger than $b^*(o,g)$, we say bias has been amplified.
For example, if $b^*(\texttt{cooking}, \texttt{woman}) = .66$, and $\tilde{b}(\texttt{cooking}, \texttt{woman}) = .84$, then the bias of \texttt{woman} toward \texttt{cooking} has been amplified.
Finally, we define the mean bias amplification as:

%To visualize the bias in the models, we propose to plot the bias scores 
%of the system predictions $\tilde{b}(o,g)$ and the labeled answers $b^*(o,g)$ 
%for each $o$.
%The plot demonstrates how 
%the correlations are enhanced or diminished by the model (see examples in the later section).

%\kw{now we have the notations, basically, we can said we want to visualize the correlation between $g$ and $o$}
%For example, in imSitu, we show the gender bias by different gender ratio for each activity in the dataset. The ratio of ``man'' of some verbs is much higher in the predictions than that in the ground.
%{\color{red}{didn't mention imSitu before, added at the cooking example.}}

%\kw{We can consider remove or simplify this section if we need space. Also, we can consider defining the bias by the difference to the tolerance margin -- this may connect better to the debias algoirhtm we desgined}
%I agree, we should probably call this section like, applified bias

%\paragraph{Quantifying the bias.} 
%We are interested in analyzing how the bias score (Eq. \eqref{eq:b_ratio})
%of the system predictions  deviates from the true distribution. 

%\begin{equation}
 %   \label{eq:bias_score}
$$\frac{1}{|O|}\sum_g \sum_{o \in \{o\in O \mid b^*(o,g) > 1/\|G\| \}} 
	\tilde{b}(o,g) - b^*(o,g). $$
%\end{equation}
This score estimates the average magnitude of bias amplification for pairs of $o$ and $g$ which exhibited bias.
%, and the magnitude how different the bias scores of  
%system predictions and gold annotations are.  

%% file: sections/method_calibrate.tex
% The debiasing algorithm will be based on the Lagrangian Relaxation by adding some corpus-wised constraints. With such algorithm we do not need to change the original inference method. In this section, we will first describe the original inference problem and then explain the constraints we add to the models.  In the end we will  provide the approach for the debiasing.
In this section, we introduce {\bf R}educing {\bf B}ias {\bf A}mplification, \alg{}, a debiasing technique for calibrating the predictions from a structured prediction model.
The intuition behind the algorithm is to inject constraints to ensure the model predictions follow the distribution observed from the training data.
For example, the constraints added to the vSRL system ensure the gender ratio of each verb in Eq.~\eqref{eq:gender_ratio} are within a given margin based on the statistics of the training data.
These constraints are applied at the corpus level, because computing gender ratio requires the predictions of all test instances.
As a result, a joint inference over test instances is required\footnote{A sufficiently large sample of test instances must be used so that bias statistics can be estimated. In this work we use the entire test set for each respective problem. }.
Solving such a giant inference problem with constraints is hard.
Therefore, we present an approximate inference algorithm based on Lagrangian relaxation. The advantages of this approach are:
%It also requires all test instances to be present at inference time, which is not practical for applications where test instances come in a streaming fashion.
\begin{itemize}
    \item Our algorithm is iterative, and at each iteration, the joint inference problem is decomposed to a per-instance basis. This can be solved by the original inference algorithm. That is, our approach works as a meta-algorithm and developers do not need to implement a new inference algorithm. 
    
    \item The approach is general and can be applied in any structured model. 
    
    \item Lagrangian relaxation guarantees the solution is optimal if the algorithm converges and all constraints are satisfied. 
\end{itemize}
In practice, it is hard to obtain a solution where all corpus-level constrains are satisfied. However, we show that the performance of the proposed approach is empirically strong.
%Interestingly, the resulting weights of the Lagrangian multipliers can be used as an offset to the feature weights, resulting in a less biased model with similar performance as the original.
We use imSitu for vSRL as a running example to explain our algorithm.

%In the following, we first provide a background of structured output prediction, then we discuss how to inject constraints to calibrate biases. The resulting constrained inference problem can be solved by various techniques. We proposed an algorithm based on Lagrangian Relaxation, in which we can use the original inference algorithm as a black-box.  We use vSRL as a running example. However, our approach is general and can be applied to any structured prediction model where the correlations between the target labels are explicitly modeled.

%We first formulate the inference problem as an Integer Linear Programming (ILP) problem.
%As the inaccessibility to the ground truth makes it impossible to set the constraints for each instance during the prediction, in this paper we will add the corpus-wise constraints.
%There can be many solutions for such problem, such as the Markov Logic Network. Here we choose to build our \alg algorithm  based on Lagrangian Relaxation  because we can just integrate the constraints to the original problem and solve it without changing the inference algorithm.

\paragraph{Structured Output Prediction}
As we mentioned in Sec. \ref{sec:general}, we assume the structured output $y \in Y$ consists of several sub-components. 
Given a test instance $i$ as an input, the inference problem is to find 
$$\arg \max_{y\in Y} \quad f_\theta (y, i),$$
where $f_\theta(y,i)$ is a scoring function based on a model $\theta$ learned from the training data. The structured output $y$ and the scoring function $f_\theta(y,i)$ can be decomposed into small components based on an independence assumption. For example, in the vSRL task, the 
output $y$ consists of two types of binary output variables $\{y_v\}$ and $\{y_{v,r}\}$.  
The variable $y_v = 1$ if and only if the activity $v$ is chosen. Similarly, $y_{v,r} = 1$ if and only if both the activity $v$ and the semantic role $r$ are assigned~\footnote{We use $r$ to refer to a combination of role and noun. For example, one possible value indicates an \texttt{agent} is a \texttt{woman}.}. 
The scoring function $f_\theta(y,i)$ is decomposed accordingly such that: $$f_\theta(y,i) = \sum_v y_v s_\theta(v,i) + \sum_{v,r}y_{v,r}s_\theta(v,r,i),$$ 
represents the overall score of an assignment, and $s_\theta(v,i)$ and $s_\theta(v,r,i)$ are the potentials of the sub-assignments. The output space $Y$ contains all feasible assignments of $y_v$ and $y_{v,r}$, which can be represented as instance-wise constraints. For example, the constraint, 
$\sum_v y_v = 1$ ensures only one activity is assigned to one image.

\paragraph{Corpus-level Constraints}
Our goal is to inject constraints to ensure the output labels follow a desired distribution. For example, we can set a constraint to ensure the gender ratio for each activity in Eq. \eqref{eq:gender_ratio} is within a given margin. 
Let $y^i = \{y^i_v\} \cup \{y^i_{v,r}\}$ be the output assignment for test instance $i$\footnote{For the sake of simplicity, we abuse the notations and use $i$ to represent both input and data index.}.
For each activity $v^*$, the constraints can be written as 
\begin{equation}\label{eq:cons_ratio} 
 b^* \!-\! \gamma \!\leq\! \frac{\sum_i y^i_{v = v^*, r \in M}}{\sum_i y^i_{v = v^*, r \in W} \!+\! \sum_i y^i_{v = v^*, r \in M}} \!\leq\! b^* + \gamma
\end{equation}
where $b^* \equiv b^*(v^*, man)$ is the desired gender ratio of an activity $v^*$, $\gamma$ is a user-specified margin. $M$ and $W$ are a set of semantic role-values representing the agent as a \texttt{man} or a \texttt{woman}, respectively. 

Note that the constraints in \eqref{eq:cons_ratio} involve all the test instances.
Therefore, it requires a joint inference over the entire test corpus.
In general, these corpus-level constraints can be represented in a form of $A\sum_i y^i - b \leq 0$, where each row in the matrix $A\in R^{l \times K}$ is the coefficients of one constraint, and $b\in R^l$.
The constrained inference problem can then be formulated as:
\begin{equation}
\begin{split}
\label{eq:objforlr}
    \max_{\{y^i\}\in \{Y^i\}} \quad& \sum_i {f_\theta(y^i, i)}, \\
      \text{s.t. } \quad& 
    A\sum_i y^i - b \leq 0,
\end{split}
\end{equation}
where $\{Y^i\}$ represents a  space spanned by possible combinations of labels for all instances. Without the corpus-level constraints, Eq. \eqref{eq:objforlr} can be optimized by maximizing each instance $i$ $$\max_{y_i\in Y^i} f_\theta (y^i, i),$$ separately.

\paragraph{Lagrangian Relaxation}
Eq. \eqref{eq:objforlr} can be solved by several combinatorial optimization methods. For example, one can represent the problem as an integer linear program and solve it using an off-the-shelf solver (e.g., Gurobi~\cite{gurobi}). However, Eq. \eqref{eq:objforlr} involves all test instances. Solving a constrained optimization problem on such a scale is difficult. Therefore, we consider relaxing the constraints and solve Eq. \eqref{eq:objforlr} using a
Lagrangian relaxation technique~\cite{Rush2012}. 
%In real applications, when given an input instance $x$, we want to get a structured prediction $y$, where the objective %function is usually defined as:
%\begin{equation} \label{eq:map}
%	y^* = \underset{y\in\mathcal{Y}}{\mathrm{argmax}}\ f(y)
%\end{equation} 
We introduce a Lagrangian multiplier $\lambda_j \geq 0$ for each corpus-level constraint. The Lagrangian is
\begin{equation}\label{eq:lag}
\begin{split}
 &L(\lambda, \{y^i\}) = \\
 &\sum_i f_\theta(y^i) - \sum_{j=1}^l \lambda_j  \left(A_j \sum_i y^i - b_j \right),
 \end{split}
\end{equation}
where all the $\lambda_j \geq 0, \forall j \in \{1,\ldots,l\}$.
% Eq~\eqref{eq:lag} can be solved by it dual problem. The dual objective is: 
% $$L(\lambda) = \max_{y \in \mathcal{Y'}} L(\lambda, y)$$
% The dual problem is to find:
% \begin{equation}\label{eq:dua}
%  \min_{\lambda \in \mathbb{R}^k} L(\lambda)
% \end{equation}
The solution of Eq. \eqref{eq:objforlr} can be obtained by the following iterative procedure:
\begin{enumerate}[label={\arabic*})]
    \item At iteration $t$, get the output solution of each instance $i$
    \begin{equation}
    \label{eq:updatey}
    y^{i,(t)} = \underset{y \in \mathcal{Y'}}{\mathrm{argmax}}\ L(\lambda^{(t-1)}, y)
    \end{equation}
    \item update the Lagrangian multipliers. 
    \begin{equation*}
     \label{eq:updatel}
     \lambda^{(t)} \!=\! \max \left(0, \lambda^{(t-1)} \!+\! \sum_i \eta(Ay^{i, (t)} -b)\right),
    \end{equation*}
\end{enumerate}
 where $\lambda^{(0)} = \boldsymbol{0}$. $\eta$ is the learning rate for updating $\lambda$. Note that with a fixed $\lambda^{(t-1)}$, Eq. \eqref{eq:updatey} can be solved using the original inference algorithms. 
 The algorithm loops until all constraints are satisfied (i.e. optimal solution achieved) or reach maximal number of iterations.

%% file: sections/experimental_setup.tex
\begin{table}[t]
\centering
    \begin{tabular}{|@{ }c@{ }|@{ }c@{ }|@{ }c@{ }|@{ }c@{ }|@{ }c@{ }|}
    \hline
    \bf Dataset  & \bf Task &  \bf Images & \bf $O$-Type & \bf $\|O\|$ \\ \hline
        imSitu  &  vSRL  & 60,000 & verb  & 212 \\ \hline
        MS-COCO    & MLC  & 25,000 & object & 66 \\ \hline
    \end{tabular}
    \caption{Statistics for the two recognition problems. In vSRL, we consider gender bias relating to verbs, while in MLC we consider the gender bias related to objects.}
    \label{tab:two_datasets}
\end{table}

In this section, we provide details about the two visual recognition tasks we evaluated for bias: visual semantic role labeling (vSRL), and multi-label classification (MLC). 
We focus on gender, defining $G = \{\texttt{man}, \texttt{woman}\}$ and focus on the \texttt{agent} role in vSRL, and any occurrence in text associated with the images in MLC.
%In each case we select a subset of the data strongly associated with humans, and build on %previous work to create a Conditional Random Field (CRF) for performing the recognition problem. 
Problem statistics are summarized in Table~\ref{tab:two_datasets}.
We also provide setup details for our calibration method.

\subsection{Visual Semantic Role Labeling}
\paragraph{Dataset}
We evaluate on imSitu~\cite{yatskar2016situation} where activity classes are drawn from verbs and roles in FrameNet~\cite{baker1998berkeley} and noun categories are drawn from WordNet~\cite{MillerBeFeGrMi90}. The original dataset includes about 125,000 images with 75,702 for training, 25,200 for developing, and 25,200 for test. However, the dataset covers many non-human oriented activities (e.g., \texttt{rearing}, \texttt{retrieving}, and \texttt{wagging}), so we filter out these verbs, resulting in 212 verbs, leaving roughly 60,000 of the original 125,000 images in the dataset. 
%For example, we filter away verbs such as \texttt{waddling} or \texttt{flapping}.

\paragraph{Model}
We build on the baseline CRF released with the data, which has been shown effective compared to a non-structured prediction baseline~\cite{yatskar2016situation}. The model decomposes the probability of a realized situation, $y$, the combination of activity, $v$, and realized frame, a set of semantic (role,noun) pairs $(e,n_e)$, given an image $i$ as :
$$ p(y |i; \theta) \propto \psi(v,i; \theta)\prod_{(e,n_e) \in R_f} \psi(v,e,n_e,i; \theta)$$
where each potential value in the CRF for subpart $x$, is computed using features $f_i$ from the VGG convolutional neural network~\cite{simonyan2014very} on an input image, as follows: $$\psi(x, i; \theta) = e^{w_x^{T}f_i + b_x},$$ where $w$ and $b$ are the parameters of an affine transformation layer. The model explicitly captures the correlation between activities and nouns in semantic roles, allowing it to learn common priors. 
We use a model pretrained on the original task with 504 verbs. 

\subsection{Multilabel Classification}
\paragraph{Dataset}
We use MS-COCO~\cite{lin2014microsoft}, a common object detection benchmark, for multi-label object classification.
The dataset contains 80 object types but does not make gender distinctions between man and woman.
We use the five associated image captions available for each image in this dataset to annotate the gender of people in the images. 
If any of the captions mention the word man or woman we mark it, removing any images that mention both genders. 
Finally, we filter any object category not strongly associated with humans by removing objects that do not occur with man or woman at least 100 times in the training set, leaving a total of 66 objects.
\paragraph{Model}
For this multi-label setting, we adapt a similar model as the structured CRF we use for vSRL. 
We decompose the joint probability of the output $y$, consisting of all object categories, $c$, and gender of the person, $g$, given an image $i$  as:
$$p(y |i ; \theta) \propto \psi(g,i; \theta)\prod_{c \in y} \psi(g,c,i; \theta)$$
where each potential value for $x$, is computed using features, $f_i$, from a pretrained ResNet-50 convolutional neural network evaluated on the image, $$\psi(x, i; \theta) = e^{w_x^{T}f_i + b_x}.$$
We trained a model using SGD with learning rate $10^{-5}$, momentum $0.9$ and weight-decay $10^{-4}$, fine tuning the initial visual network, for $50$ epochs.

\subsection{Calibration}
The inference problems for both models are: $$\arg\max_{y\in Y} f_\theta(y, i) = \log p(y|i; \theta).$$ 

We use the algorithm  in Sec. \eqref{sec:calibration} to calibrate the predictions using model $\theta$.
Our calibration tries to enforce gender statistics derived from the training set of corpus applicable for each recognition problem. 
For all experiments, we try to match gender ratios on the test set within a margin of $.05$ of their value on the training set. 
While we do adjust the output on the test set, we never use the ground truth on the test set and instead working from the assumption that it should be similarly distributed as the training set.
When running the debiasing algorithm, we set $\eta = 10^{-1}$ and optimize for $100$ iterations.

%% file: sections/results_bias.tex
\begin{figure*}[t]
    \centering
    \vspace{-10pt}
    \hspace{-25pt}
    % \hspace*{-2cm}
    ~ %add desired spacing between images, e. g. ~, \quad, \qquad, \hfill etc. 
            %\hspace{-20pt}
    \begin{subfigure}[b]{0.48\textwidth}
        \includegraphics[width=1.15\linewidth]{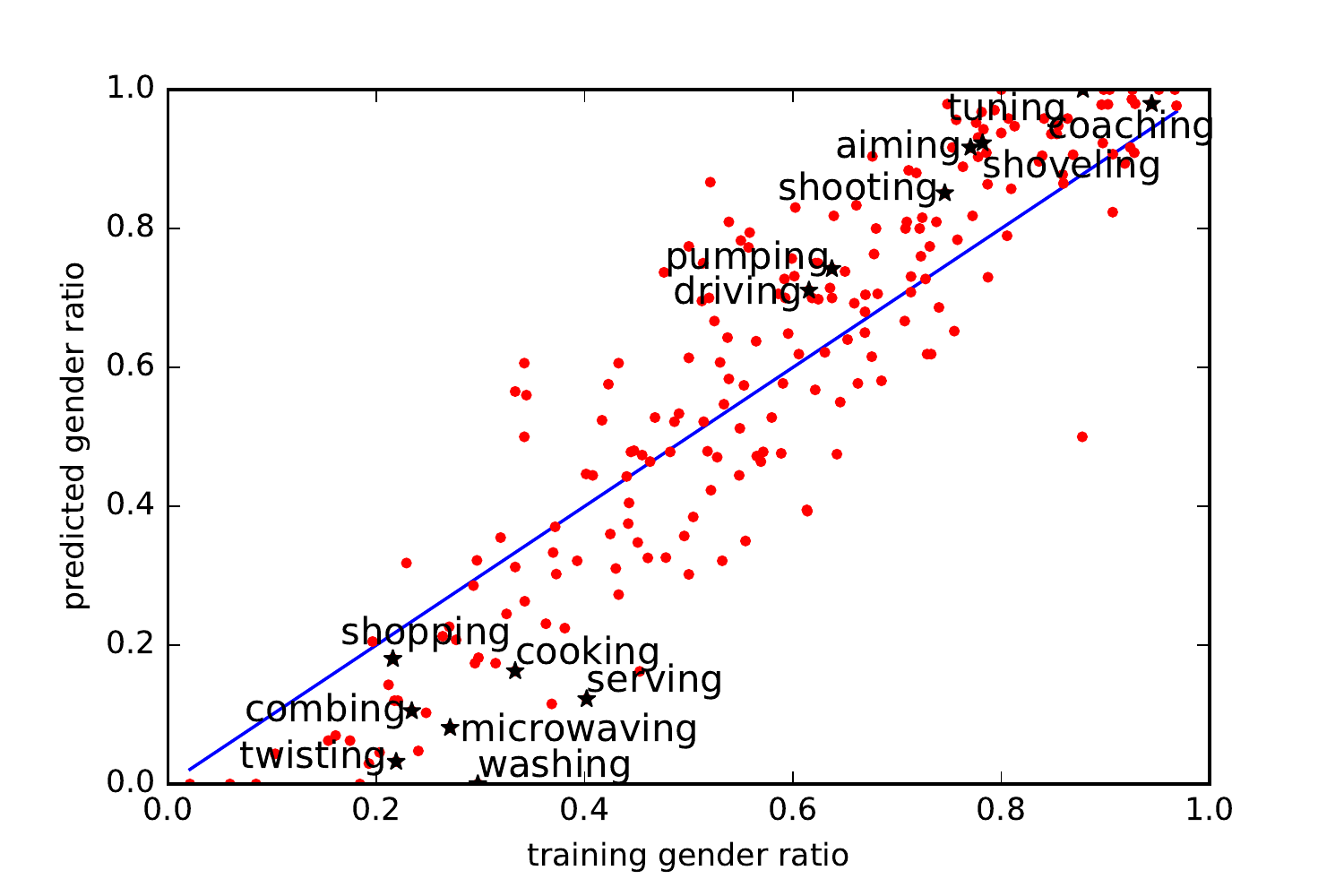}
        \caption{Bias analysis on imSitu vSRL}
        \label{fig:biased_verb_dev}
    \end{subfigure}
   % \hspace{-16pt}
  %  \begin{subfigure}[b]{0.26\textwidth}
  %      \includegraphics[width=\linewidth]{figures/gender_ratio_train.pdf}
  %      \caption{Gender ratio on training. }
  %      \label{fig:biased_verb_train}
  %  \end{subfigure}
    %~ %add desired spacing between images, e. g. ~, \quad, \qquad, \hfill etc. 
  %        \hspace{-20pt}
  %  \begin{subfigure}[b]{0.26\textwidth}
  %      \includegraphics[width=\linewidth]{figures/meanVerbRatio_for_man_dev.pdf}
  %      \caption{Verb distribution of ``men''.}
  %      \label{fig:verb_ratio}
  %  \end{subfigure}
  %        \hspace{-20pt}
    ~ %add desired spacing between images, e. g. ~, \quad, \qquad, \hfill etc. 
  %  \begin{subfigure}[b]{0.42\textwidth}
  %      \includegraphics[width=\linewidth]{figures/bias_score_each_verb_oridev_imSitu.pdf}
  %      \caption{Bias amplification across verb by confidence threshold, averaged across verbs.
  %       Values near zero indicate no bias amplification.
        %As the system becomes less confident, it amplifies bias more. Even at high confidence thresholds, bias amplification persists.
  %      }
   % \label{fig:bias_score_imSitu}
%    \end{subfigure}
   \begin{subfigure}[b]{0.48\textwidth}
        \includegraphics[width=1.15\linewidth]{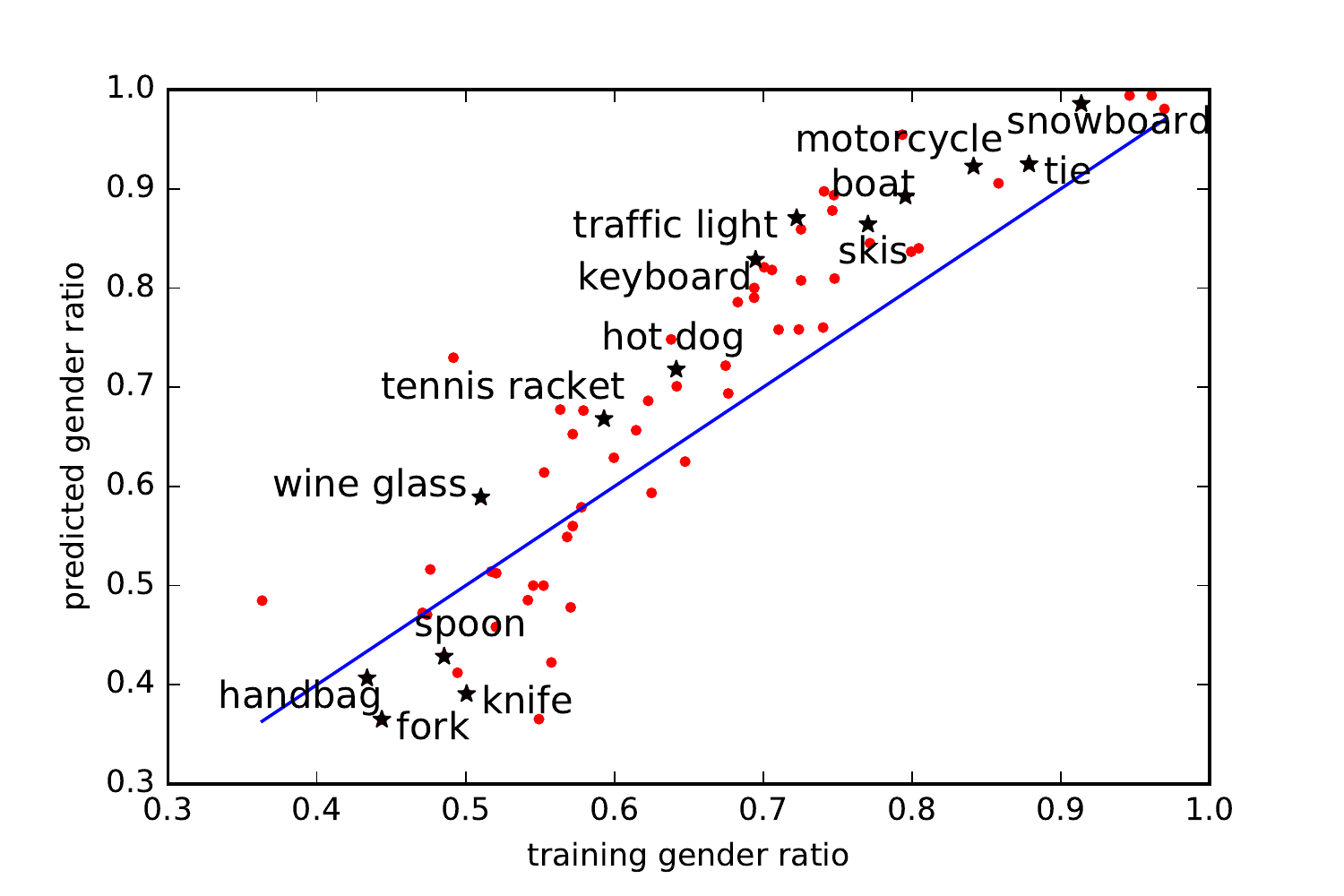}
        % \caption{Gender bias of MS-COCO objects toward \texttt{man} in training set versus bias on a predicted development set. Values near zero indicate bias}
        \caption{Bias analysis on MS-COCO MLC}
        \label{fig:gen_dev_coco}
    \end{subfigure}
    \caption{Gender bias analysis of imSitu vSRL and MS-COCO MLC. (a) gender bias of verbs toward \texttt{man} in the training set versus bias on a predicted development set. (b) gender bias of nouns toward \texttt{man} in the training set versus bias on the predicted development set. Values near zero indicate bias toward \texttt{woman} while values near $0.5$ indicate unbiased variables. Across both dataset, there is significant bias toward males, and significant bias amplification after training on biased training data.}
    \label{fig:imSitu_results}
\end{figure*}

%\begin{figure*}[t]
%    \centering
    % \hspace*{-2cm}
    ~ %add desired spacing between images, e. g. ~, \quad, \qquad, \hfill etc. 
%    \begin{subfigure}[b]{0.42\textwidth}
%        \includegraphics[width=\linewidth]{figures/gen_ratio_obj_coco_test.pdf}
%        \caption{Gender bias of MS-COCO objects toward \texttt{man} in training set versus bias on a predicted dev set.}
%        \label{fig:gen_dev_coco}
%    \end{subfigure}
   % ~
   %\hspace{-0pt}
%     \begin{subfigure}[b]{0.31\textwidth}
 %       \includegraphics[width=\linewidth]{figures/gen_ratio_obj_coco_train.pdf}
  %      \caption{Gender ratio on train. }
   %     \label{fig:gen_train_coco}
    %\end{subfigure}
    ~ %add desired spacing between images, e. g. ~, \quad, \qquad, \hfill etc. 
%    \begin{subfigure}[b]{0.42\textwidth}
%        \includegraphics[width=\linewidth]{figures/gen_ratio_super_coco.pdf}
%        \caption{Gender bias of  MS-COCO object toward man, grouped by object super-categories.}
%        \label{fig:gen_sup_coco}
%    \end{subfigure}
%    \caption{
%    Gender bias analysis in MS-COCO MLC. The graph on the left indicates near universal bias of categories toward \texttt{men}, with bias amplification in the development set after training. The graph on the right indicates bias amplification across all super-categories in MS-COCO, especially pronounced for the \texttt{outdoor} super category.} 
%    \label{fig:coco_results}
%\end{figure*}
In this section, we use the approaches outlined in Section \ref{sec:general} to quantify the bias and bias amplification in the vSRL and the MLC tasks.
\subsection{Visual Semantic Role Labeling}
\paragraph{imSitu is gender biased}
In Figure~\ref{fig:biased_verb_dev}, along the x-axis, we show the male favoring bias of imSitu verbs. 
Overall, the dataset is heavily biased toward male agents, with 64.6\% of verbs favoring a male agent by an average bias of $0.707$ (roughly 3:1 male).
Nearly half of verbs are extremely biased in the male or female direction: 46.95\% of verbs favor a gender with a bias of at least $0.7$.\footnote{In this gender binary, bias toward \texttt{woman} is $1 - $ the bias toward \texttt{man}}
Figure~\ref{fig:biased_verb_dev} contains several activity labels revealing problematic biases.
For example, \texttt{shopping}, \texttt{microwaving} and \texttt{washing} are biased toward a female \texttt{agent}.
Furthermore, several verbs such as \texttt{driving}, \texttt{shooting}, and \texttt{coaching} are heavily biased toward a male \texttt{agent}.

\paragraph{Training on imSitu amplifies bias}
In Figure~\ref{fig:biased_verb_dev}, along the y-axis, we show the ratio of male agents (\% of total people) in predictions on an unseen development set.
The mean bias amplification in the development set is high, $0.050$ on average, with $45.75\%$ of verbs exhibiting amplification.
Biased verbs tend to have stronger amplification: verbs with training bias over $0.7$ in either the male or female direction have a mean amplification of $0.072$.
Several already problematic biases have gotten much worse.
For example, \texttt{serving}, only had a small bias toward females in the training set, $0.402$, is now heavily biased toward females, $0.122$.
The verb \texttt{tuning}, originally heavily biased toward males, $0.878$, now has exclusively male agents.

\subsection{Multilabel Classification}
\paragraph{MS-COCO is gender biased}
In Figure~\ref{fig:gen_dev_coco} along the x-axis, similarly to imSitu, we analyze bias of objects in MS-COCO with respect to males. 
MS-COCO is even more heavily biased toward men than imSitu, with $86.6\%$ of objects biased toward men, but with smaller average magnitude, $0.65$.
One third of the nouns are extremely biased toward males, 37.9\% of nouns favor men with a bias of at least $0.7$.
Some problematic examples include kitchen objects such as \texttt{knife}, \texttt{fork}, or \texttt{spoon} being more biased toward woman.
Outdoor recreation related objects such \texttt{tennis racket}, \texttt{snowboard} and \texttt{boat} tend to be more biased toward men.

\paragraph{Training on MS-COCO amplifies bias}
In Figure~\ref{fig:gen_dev_coco}, along the y-axis, we show the ratio of man {(\% of both gender)} in predictions on an unseen development set.
The mean bias amplification across all objects is $0.036$, with $65.67\%$ of nouns exhibiting amplification.
Larger training bias again tended to indicate higher bias amplification: biased objects with training bias over $0.7$ had mean amplification of $0.081$.
Again, several problematic biases have now been amplified.
For example, kitchen categories already biased toward females such as \texttt{knife}, \texttt{fork} and \texttt{spoon} have all been amplified.
Technology oriented categories initially biased toward men such as \texttt{keyboard} and \texttt{mouse} have each increased their bias toward males by over $0.100$.

\subsection{Discussion}
We confirmed our hypothesis that (a) both the imSitu and MS-COCO datasets, gathered from the web, are heavily gender biased and that (b) models trained to perform prediction on these datasets amplify the existing gender bias when evaluated on development data.
Furthermore, across both datasets, we showed that the degree of bias amplification was related to the size of the initial bias, with highly biased object and verb categories exhibiting more bias amplification.
Our results demonstrate that care needs be taken in deploying such uncalibrated systems otherwise they could not only reinforce existing social bias but actually make them worse.

%% file: sections/results_calibration.tex
%\section{Experiments}
We test our methods for reducing bias amplification in two problem settings: visual semantic role labeling in the imSitu dataset (vSRL) and multilabel image classification in MS-COCO (MLC). 
In all settings we derive corpus constraints using the training set and then run our calibration method in batch on either the development or testing set. 
Our results are summarized in Table~\ref{tab:results} and Figure~\ref{fig:results}.
%using two different settings:  \textit{batch} setting and \textit{online} setting. 
%In batch setting, we have access to all the test instances. 
%Therefore, we can calibrate the gender ratio on them. 
%We show this result on the dev set. In online setting, we only can access one instance at a time. 
%Therefore, we cannot add corpus-wise constraints. 
%However, we show that we can still remove the amplified biases in such test instance by applying the offset we obtained from the batch setting.
% We get the lambdas using the training and dev dataset and then adopt it to the test dataset. Besides trying to remove gender bias, we also test our \alg again on the verb bias issue, because we notice that for some verbs like ``fixing'' are more likely to appear when the system detects there is a man  in the image.
\subsection{Visual Semantic Role Labeling}
\begin{figure*}[!h]
    \centering
    % \hspace{-30pt}
       \begin{subfigure}[b]{0.49\textwidth}
        \includegraphics[width=1.1\linewidth]{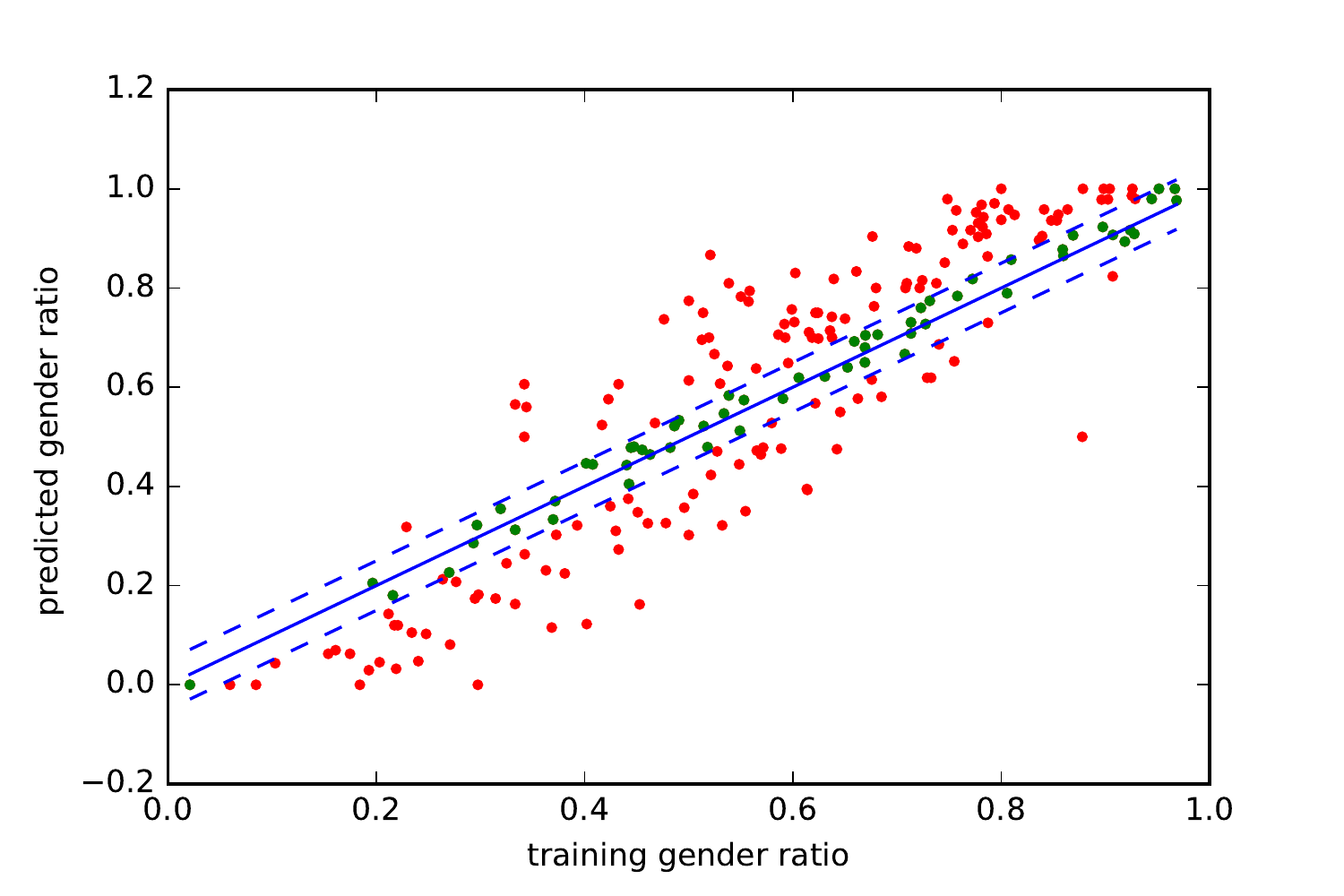}
        \caption{Bias analysis on imSitu vSRL without \alg}
        \label{fig:gender_ratio_beforeLR}
    \end{subfigure}
    %   \hspace{-16pt}
    \begin{subfigure}[b]{0.49\textwidth}
        \includegraphics[width=1.1\linewidth]{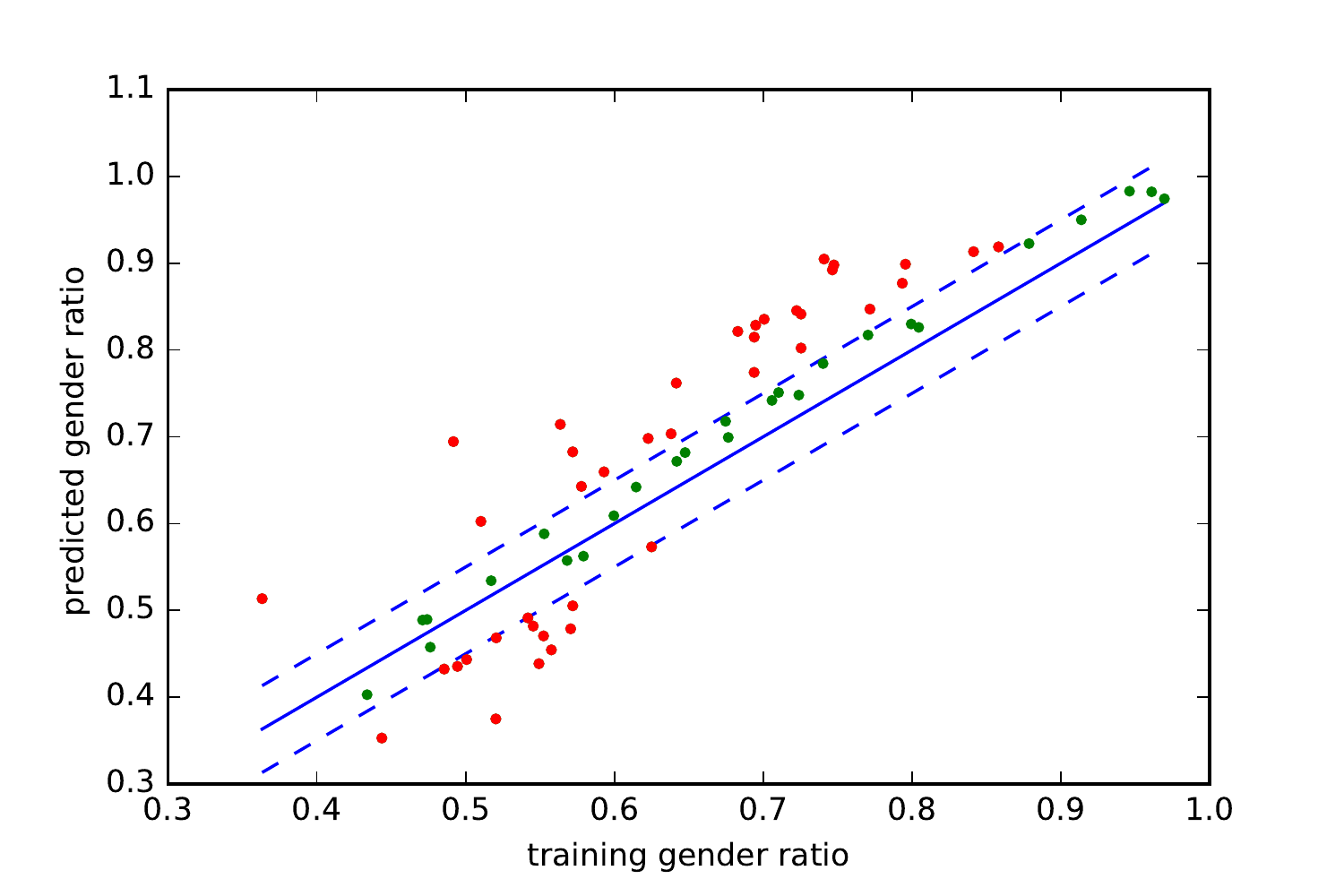}
        \caption{Bias analysis on MS-COCO MLC without \alg}
        \label{fig:coco_gender_ratio_beforeLR}
    \end{subfigure}
      \vspace{10pt}
    %   \hspace{-16pt}    
      \begin{subfigure}[b]{0.49\textwidth}
        \includegraphics[width=1.1\linewidth]{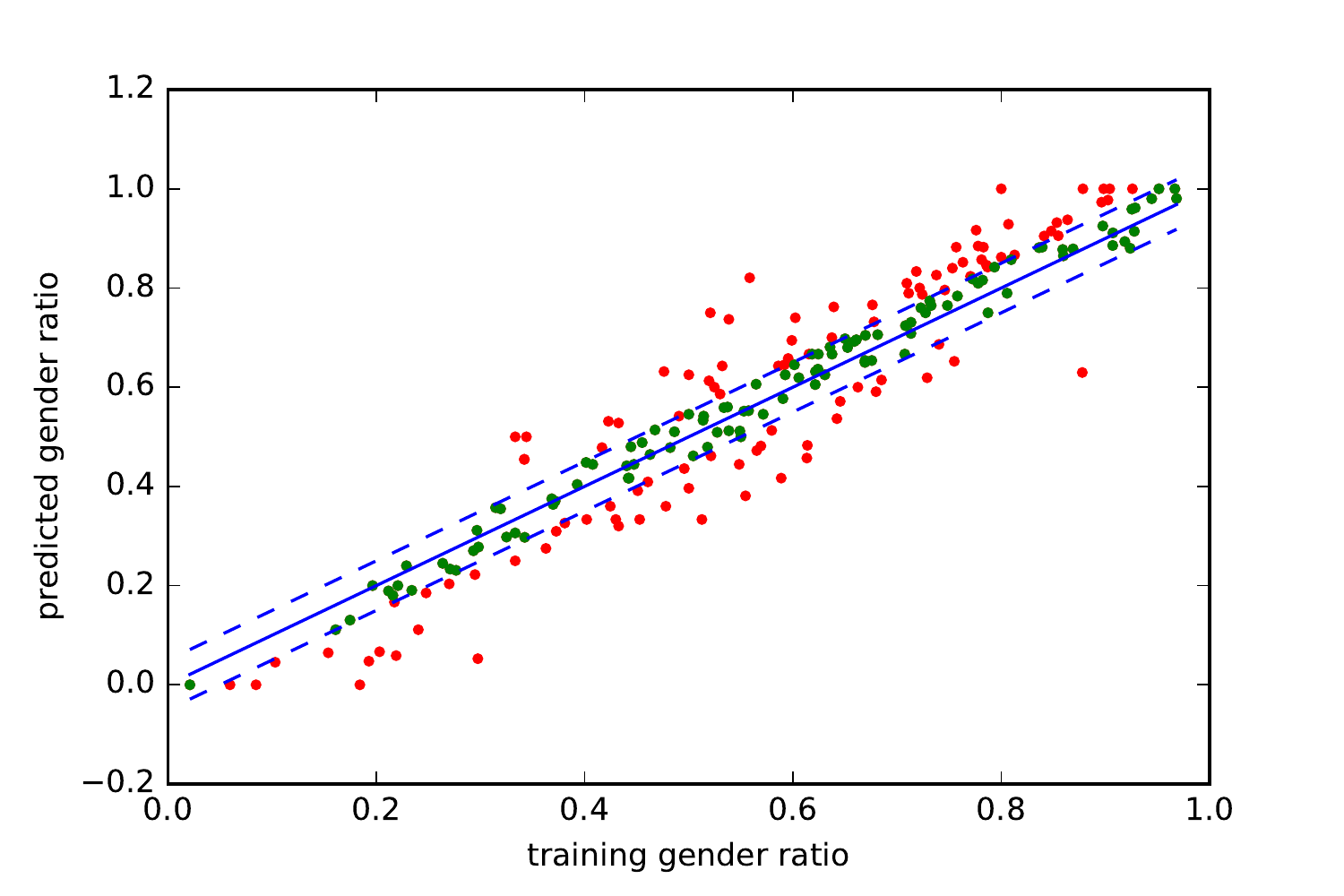}
        \caption{Bias analysis on imSitu vSRL with \alg }
        \label{fig:gender_ratio_afterLR}
    \end{subfigure}
    \begin{subfigure}[b]{0.49\textwidth}
        \includegraphics[width=1.1\linewidth]{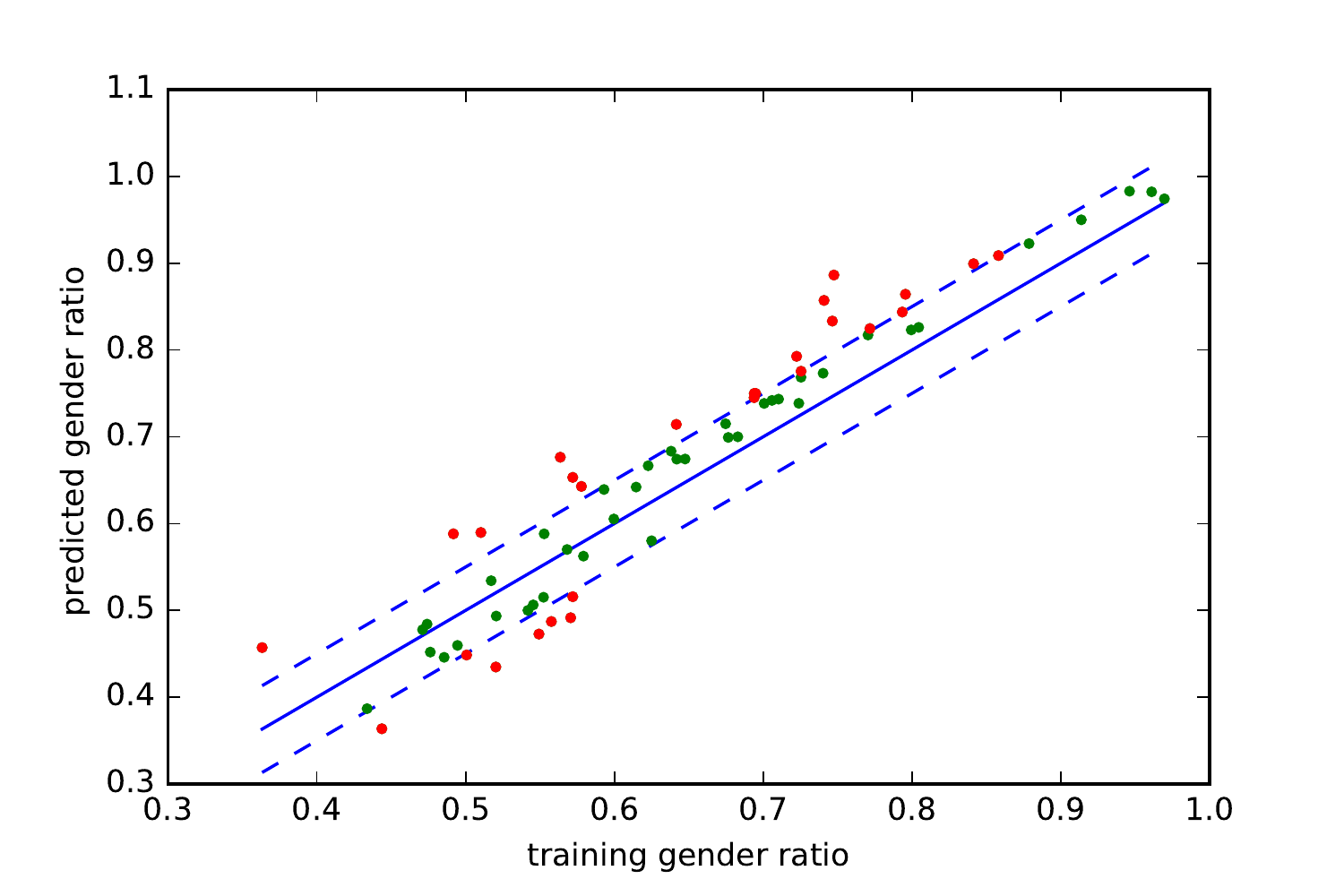}
        \caption{Bias analysis on MS-COCO MLC with \alg }
        \label{fig:coco_gender_ratio_afterLR}
    \end{subfigure}
      \vspace{10pt}
    \begin{subfigure}[b]{0.49\textwidth}
        \includegraphics[width=1.1\linewidth]{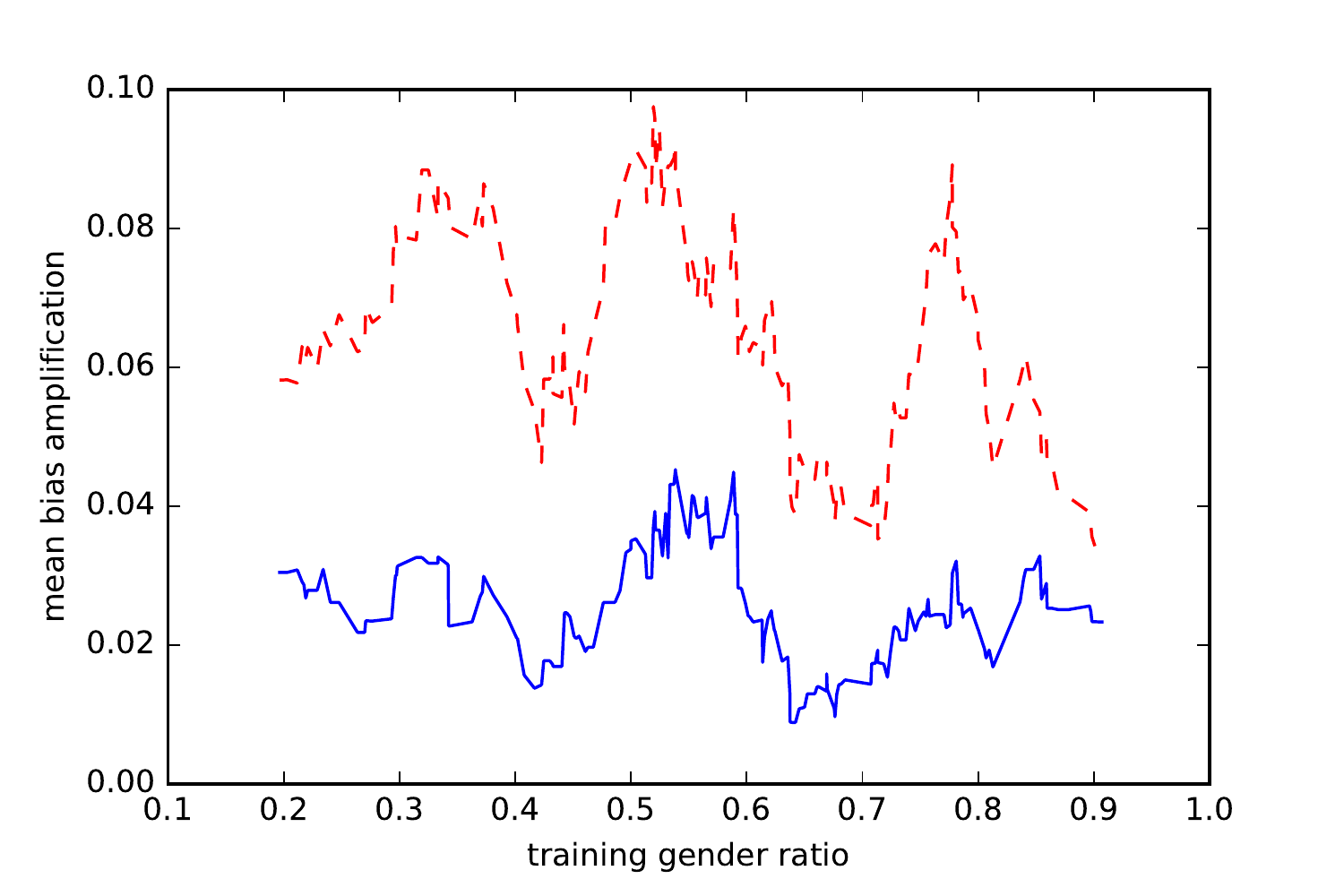}
        \caption{Bias in vSRL with (blue) / without (red) \alg}
        \label{fig:mean_gender_ratio_dis}
    \end{subfigure}    
    \begin{subfigure}[b]{0.49\textwidth}
        \includegraphics[width=1.1\linewidth]{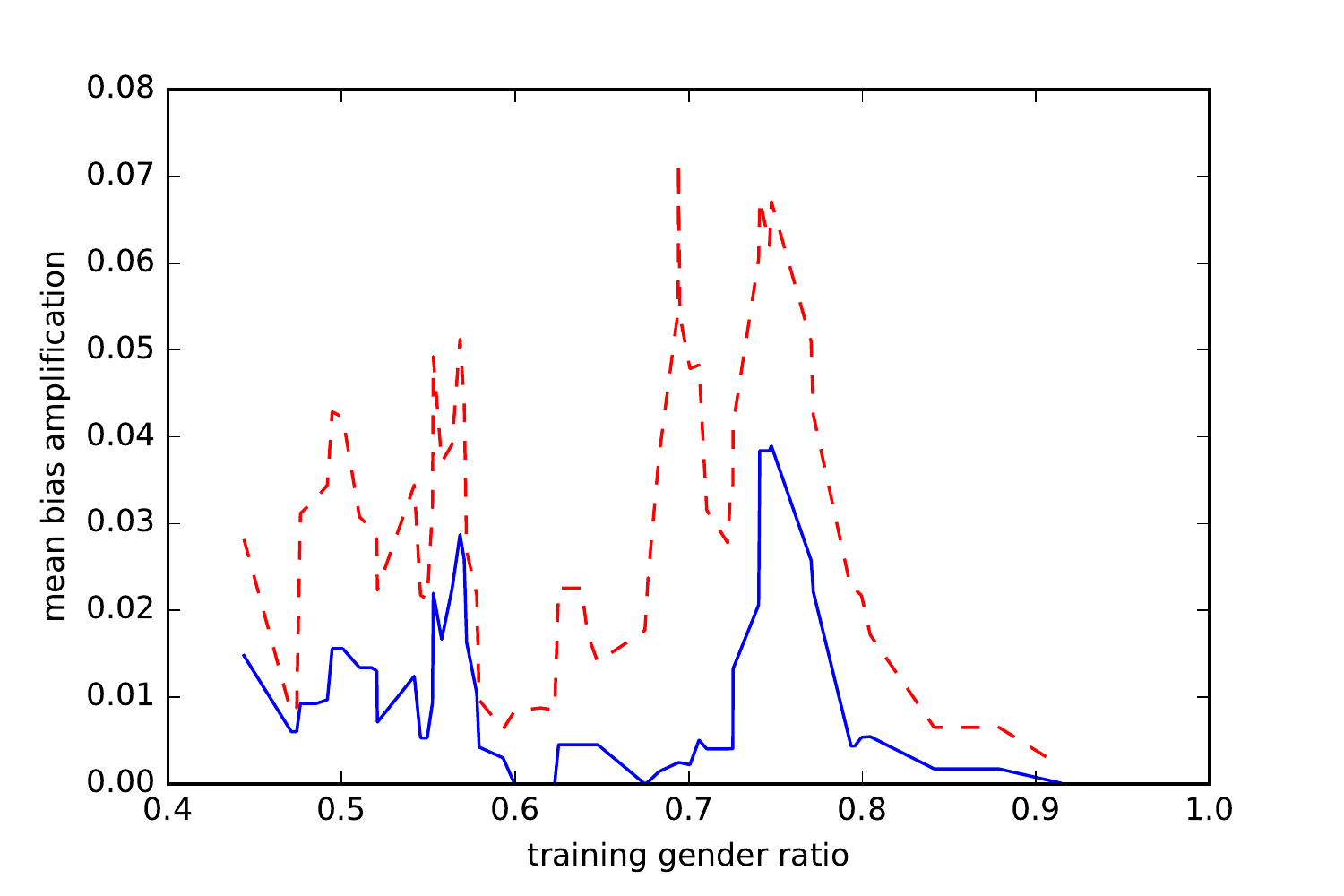}
        \caption{Bias in MLC with (blue) / without (red) \alg}
        \label{fig:coco_mean_gender_ratio_dis}
    \end{subfigure}
    \caption{
    Results of reducing bias amplification using \alg on imSitu vSRL and MS-COCO MLC.
    Figures 3(a)-(d) show initial training set bias along the x-axis and development set bias along the y-axis. 
    Dotted blue lines indicate the $0.05$ margin used in \alg, with points violating the margin shown in red while points meeting the margin are shown in green. 
    Across both settings adding \alg significantly reduces the number of violations, and reduces the bias amplification significantly. 
    Figures 3(e)-(f) demonstrate bias amplification as a function of training bias, with and without \alg. 
    Across all initial training biases, \alg  is able to reduce the bias amplification.
    } 
    \label{fig:results}
\end{figure*}

Our quantitative results are summarized in the first two sections of Table~\ref{tab:results}.
On the development set, the number of verbs whose bias exceed the original bias by over 5\% decreases 30.5\% (Viol.).
Overall, we are able to significantly reduce bias amplification in vSRL by 52\% on the development set (Amp. bias).
We evaluate the underlying recognition performance using the standard measure in vSRL:  top-1 semantic role accuracy, which tests how often the correct verb was predicted and the noun value was correctly assigned to a semantic role.
Our calibration method results in a negligible decrease in performance (Perf.).
In Figure~\ref{fig:gender_ratio_afterLR} we can see that the overall distance to the training set distribution after applying \alg{} decreased significantly, over 39\%.

\begin{table}[t]
\centering
    \begin{tabular}{|l|c|c|c|}
 
        \hline
         \bf Method & \bf Viol. & \bf Amp. bias & \bf Perf. (\%) \\
         \hline
          \multicolumn{4}{|c|}{vSRL: Development Set} \\
          \hline
        CRF  &  154   & 0.050  & 24.07 \\
         \hline
        CRF + \alg &  107 & 0.024  & 23.97\\
        \hline
        \multicolumn{4}{|c|}{vSRL: Test Set} \\
        \hline
        CRF  &  149   & 0.042 & 24.14 \\
         \hline
        CRF + \alg &  102 & 0.025 & 24.01\\
        \hline
          \multicolumn{4}{|c|}{MLC: Development Set} \\
          \hline
     CRF  &  40   & 0.032  & 45.27 \\
         \hline
        CRF + \alg &  24 & 0.022  & 45.19\\
        \hline
    \multicolumn{4}{|c|}{MLC: Test Set} \\
         \hline
          CRF  &  38   & 0.040  &  45.40\\
         \hline
        CRF + \alg &  16 & 0.021  & 45.38\\
        \hline
    \end{tabular}
    \caption{
    Number of violated constraints, mean amplified bias, and test performance before and after calibration using \alg{}. 
    The test performances of vSRL and MLC are measured by top-1 semantic role accuracy and  top-1 mean average precision, respectively.}
    \label{tab:results}
\end{table}

Figure~\ref{fig:mean_gender_ratio_dis} demonstrates that across all initial training bias, \alg{} is able to reduce bias amplification. 
In general, \alg{} struggles to remove bias amplification in areas of low initial training bias, likely because bias is encoded in image statistics and cannot be removed as effectively with an image agnostic adjustment.
Results on the test set support our development set results: we decrease bias amplification by 40.5\% (Amp. bias).
\subsection{Multilabel Classification}
Our quantitative results on MS-COCO RBA are summarized in the last two sections of Table~\ref{tab:results}. 
Similarly to vSRL, we are able to reduce the number of objects whose bias exceeds the original training bias by 5\%, by 40\% (Viol.).
Bias amplification was reduced by 31.3\% on the development set (Amp. bias). 
The underlying recognition system was evaluated by the standard measure: top-1 mean average precision, the precision averaged across object categories.
Our calibration method results in a negligible loss in performance.
In Figure~\ref{fig:coco_gender_ratio_afterLR}, we demonstrate that we substantially reduce the distance between training bias and bias in the development set.
Finally, in Figure~\ref{fig:coco_mean_gender_ratio_dis} we demonstrate that we decrease bias amplification for all initial training bias settings. 
Results on the test set support our development results: we decrease bias amplification by 47.5\% (Amp. bias).
\subsection{Discussion}
We have demonstrated that \alg{} can significantly reduce bias amplification.
While were not able to remove all amplification, we have made significant progress with little or no loss in underlying recognition performance. 
Across both problems, \alg{} was able to reduce bias amplification at all initial values of training bias.

%% file: sections/conclusion.tex
Structured prediction models can leverage correlations that allow them to make correct predictions even with very little underlying evidence.
%between output variables, which can potentially result in bias and, as we showed, amplified bias.
Yet such models risk potentially leveraging social bias in their training data.
In this paper, we presented a general framework for visualizing and quantifying biases in such models and proposed \alg to calibrate their predictions under two different settings.
Taking gender bias as an example, our analysis demonstrates that conditional random fields can amplify social bias from data while our approach \alg{} can help to reduce the bias.

Our work is the first to demonstrate structured prediction models amplify bias and the first to propose methods for reducing this effect but significant avenues for future work remain.
While \alg{} can be applied to any structured predictor, it is unclear whether different predictors amplify bias more or less.
Furthermore, we presented only one method for measuring bias. 
More extensive analysis could explore the interaction among predictor, bias measurement, and bias de-amplification method. 
Future work also includes applying bias reducing methods in other structured domains, such as pronoun reference  resolution~\cite{mitkov2014anaphora}.

\paragraph{Acknowledgement}
This work was supported in part by National Science Foundation Grant IIS-1657193 and two NVIDIA Hardware Grants.